\documentclass{article}

\PassOptionsToPackage{numbers, compress}{natbib}

\usepackage[preprint]{neurips_2026}
\usepackage{wrapfig}
\usepackage{booktabs}
\usepackage{multirow}
\usepackage{graphicx} 
\usepackage{booktabs} 
\usepackage{makecell}
\usepackage{xcolor}
\usepackage{colortbl}
\usepackage{pifont}
\usepackage{array}
\usepackage{amssymb}
\newcommand{\up}[1]{\textcolor{green}{\small$\uparrow$#1}}
\newcommand{\down}[1]{\textcolor{red}{\small$\downarrow$#1}}

\usepackage[utf8]{inputenc} 
\usepackage[T1]{fontenc}    
\usepackage{url}            
\usepackage{booktabs}       
\usepackage{amsfonts}       
\usepackage{nicefrac}       
\usepackage{microtype}      
\usepackage{xcolor}         
\usepackage{graphicx}
\usepackage[table]{xcolor}
\usepackage{amsmath}

\definecolor{linkcolor}{RGB}{255,0,0}
\definecolor{urlcolor}{RGB}{255,105,180}
\definecolor{citecolor}{RGB}{66,168,235}
\usepackage[pagebackref=true,breaklinks=true,letterpaper=true,colorlinks,allcolors=blue,bookmarks=false]{hyperref}

\title{PixelEyes: Decoupling Perception and Reasoning for Pinpoint Visual Evidence Seeking}

\author{
Dengxian Gong$^{1}$\textsuperscript{*}~\quad
Yuanzheng Wu$^{1}$\textsuperscript{*}~\quad
Haobo Yuan$^{2}$\quad
Zhengdong Hu$^{3}$\quad
Tao Zhang$^{1}$~\quad \\
\textbf{Yikang Zhou}$^{1}$~\quad 
\textbf{Shihao Chen}$^{1}$~\quad
\textbf{Quanzhu Niu}$^{1}$~\quad
\textbf{Kai Wang}$^{4}$~\quad
\textbf{Jason Li}$^{5}$~\quad \\
\textbf{Haochen Wang}$^{6}$~\quad 
\textbf{Lu Qi}$^{1}$\textsuperscript{\textdagger}\quad
\textbf{Shunping Ji}$^{1}$\textsuperscript{\textdagger}\quad
\textbf{Ming-Hsuan Yang}$^{2}$\\
$^1$Wuhan University\quad
$^2$UC Merced\quad
$^3$UTS\quad
$^4$NUS\quad
$^5$NTU\quad
$^6$CASIA\quad \\
{\tt\small \textsuperscript{*}Equal contribution, {\textdagger}Corresponding authors}\\
{\tt\small \{gooodx,jishunping\}@whu.edu.cn}\\
{\small{Project page: \url{https://godx-7.github.io/PixelEyesSite/}}}
}

\begin{document}

\maketitle

\begin{figure}[!ht]
    \centering 
    \includegraphics[width=1\textwidth]{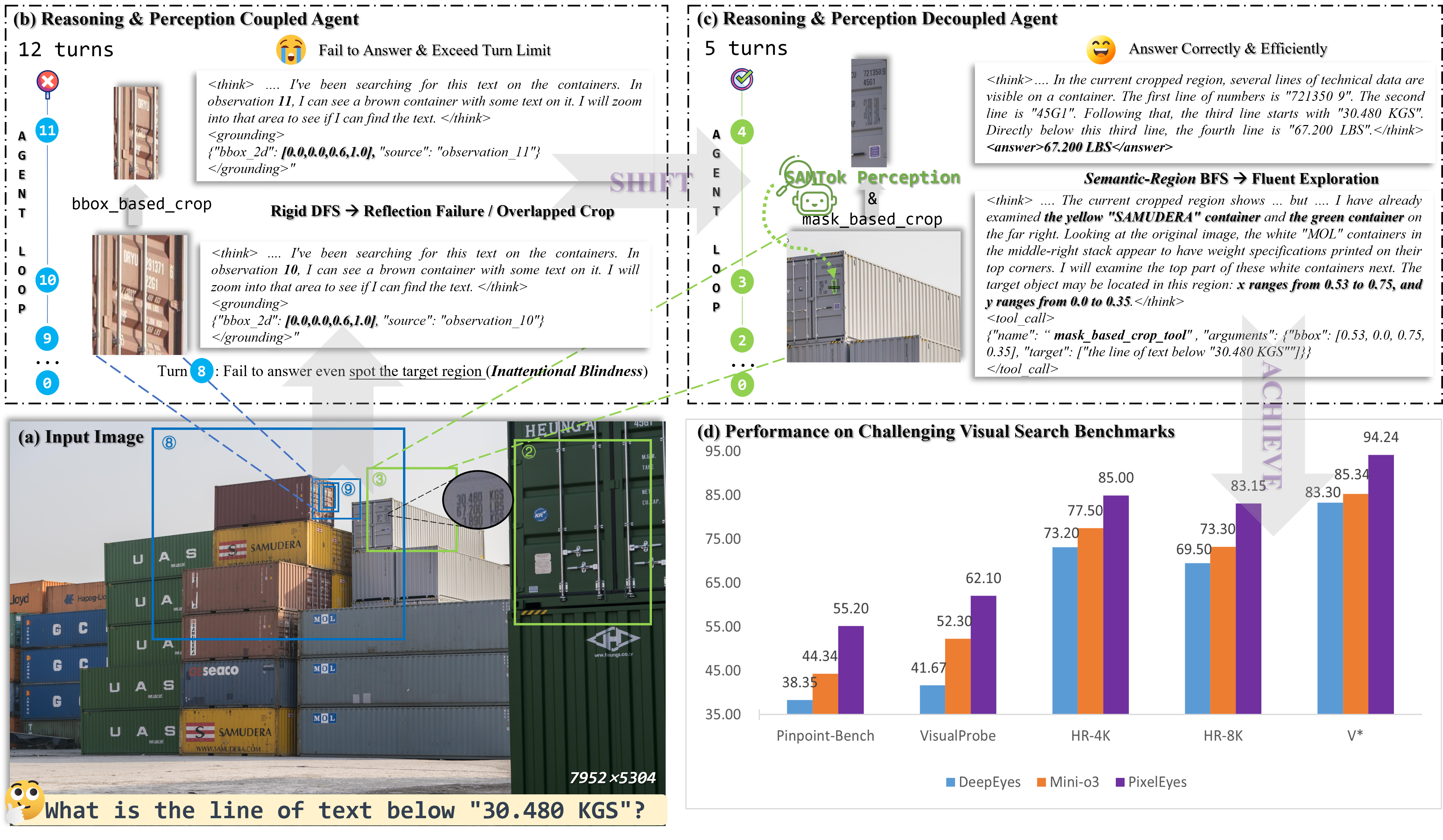}
    \caption{ 
    \textbf{Paradigm Comparison for Active Visual Search}. 
    (a) A challenging instance-anchored visual query. (b) Coupled Agent (Baseline): Relying on coarse bounding boxes, existing models suffer from "inattentional blindness" (spotting the correct region but failing to recognize the target) and fall into rigid, inefficient deep-search loops, eventually exhausting the turn limit. (c) Decoupled Agent (PixelEyes): By employing the SAMTok tool for precise, mask-guided cropping and adopting a Semantic-Region Breadth-First Search (BFS) exploration strategy, our framework eliminates background distractors and efficiently locates the target in just 5 turns. (d) Consequently, PixelEyes achieves state-of-the-art performance across multiple rigorous visual search benchmarks, significantly outperforming existing methods.
    }
    \label{fig:teaser}
\end{figure}

\begin{abstract}
This paper explores multi-turn visual reasoning and observes that MLLMs repeatedly fail to localize the target, leading to long, redundant trajectories.
We attribute this failure to the entanglement of reasoning and perception within a single model, the MLLM reasons and localizes simultaneously, and inaccurate localization triggers additional reasoning turns that bloat the trajectory.
To solve this problem, we propose \textbf{PixelEyes}, a multi-turn visual reasoning agent that explicitly decouples reasoning from perception, \emph{i.e.,} the reasoner decides \emph{what to look for}, while a specialized perception tool answers \emph{where it is}.
Specifically, PixelEyes introduces 1) \textbf{Mask-guided Visual Search}. A referring segmentation model is invoked to provide mask-precise localization, freeing the reasoner from the need to compensate for imprecise grounding. 
2) \textbf{Semantic-region Breadth-first Search (BFS)}. 
To eliminate redundant loops caused by repeatedly cropping incorrect sub-regions, we organize exploration as a breadth-first search over semantic regions.
To internalize these capabilities, we construct the \textbf{PixelEyes-6K} dataset by resynthesizing expert trajectories from existing data. This explicitly embeds our mask-guided search and BFS logic into the model. 
We further introduce \textbf{Pinpoint-Bench}, a zero-hint visual search benchmark, \emph{i.e.,} no location cues are provided in the question, with instance-level masks and bounding boxes that separate localization failures from reasoning failures, enabling fine-grained analysis of failure modes such as inattentional blindness. 
Recent state-of-the-art MLLMs and visual reasoning agents leave large headroom on Pinpoint-Bench, demonstrating its quality and difficulty.
Code and models are open-sourced.
\end{abstract}

\section{Introduction}
\label{sec:intro}

Vision-language models are moving from passive observers to active visual reasoners that crop, zoom, and re-examine an image to gather evidence -- a paradigm popularized by OpenAI~o3~\cite{openai2025o3o4mini} and now widely studied as ``Thinking with Images''~\cite{su2025thinking}.
The setting is challenging in practice: decisive evidence often lies in objects occupying less than $1\%$ of a high-resolution image, so an agent must locate a needle-in-a-haystack target and then reason about its content within a bounded turn budget.

Existing methods~\cite{su2025openthinkimg, wu2025vtool, FCOS, hu2024visual, shen2025zoomeye, mini_o3} pursue this with a single model that is asked to perform both jobs at once: fine-grained region-level perception \emph{and} general reasoning over the cropped evidence.
This coupling is uncomfortable. The same model is consistently weaker at grounding than perception-oriented specialists~\cite{lai2024lisa,zhang2024omg} and consistently weaker at reasoning than strong general-purpose VLMs~\cite{qwen3,internvl35,kimi_k25,gemini,gpt4o}.
Two failure modes follow.
First, weak perception leads to long trajectories of blind crops when the agent fails to localize the correct region.
Second, even when the correct region is cropped, degraded reasoning can cause \emph{inattentional blindness}: the agent sees the target but fails to recognize it.
We use ``inattentional blindness'' throughout the paper to denote precisely this gap between visiting and answering, and quantify it directly in our benchmark.

We propose \textbf{PixelEyes}, an agent that decouples perception from reasoning.
A general-purpose VLM~\cite{qwen25vl,qwen3,deepseekvl2,cogvlm2,glm45v,claude37} decides \emph{what} to look for; an external referring-segmentation tool, SAMTok~\cite{samtok}, answers \emph{where} it is by returning a pixel-level mask rather than a coarse bounding box~\cite{molmo2,rex-omni,guo2025seed1}.
Two further mechanisms keep the trajectory short.
(i) \emph{Semantic-Region BFS}: the reasoner anchors every coordinate in the original image and proposes a new low-IoU region whenever SAMTok fails to ground the target, expanding sibling regions before descending into any one of them.
(ii) \emph{Switchable Tool Use}: when mask grounding is ill-defined (charts, maps, dense text), the agent falls back to a plain bounding-box crop.

To internalize this behavior we synthesize \textbf{PixelEyes-6K} -- $5.8$K expert trajectories produced by augmenting Gemini-3-Flash~\cite{gemini} with the same \texttt{mask\_based\_crop} tool and rolling out closed-loop interactions on existing image-question pairs.
We retain only trajectories that reach a correct answer and use them as supervised fine-tuning data for Qwen-3-VL~\cite{qwen3}; reinforcement learning with vanilla GRPO further sharpens the policy.
Training the structure of search, rather than scaling up the number of turns, turns out to be a substantially stronger lever (Sec.~\ref {sec:exp}).

Existing benchmarks make this hard to evaluate cleanly. V*~\cite{vstar} and HR-Bench~\cite{HR_bench} are saturating; TreeBench~\cite{Treebench} and MME-RealWorld~\cite{MME} emphasize reasoning over search; VisualProbe~\cite{mini_o3} has appropriate difficulty but no spatial annotations, so localization failures cannot be separated from reasoning failures.
We introduce \textbf{Pinpoint-Bench}: 433 human-annotated samples on ultra-high-resolution images, with target masks averaging only $0.07\%$ of the image area (Tab.~\ref{tab:dataset_comparison}).
Queries follow a strict zero-hint protocol -- no spatial cues, no macro-anchors -- and answers admit multi-alias matching to absorb linguistic ambiguity.
Beyond accuracy, Pinpoint-Bench reports \emph{Localization Success Rate} (LSR), which marks a trial successful if any crop covers the target, and \emph{Turn-to-Answer Efficiency} (TAE), which normalizes accuracy by interaction turns.
The gap between LSR and accuracy quantifies inattentional blindness directly.

Our contributions are: (1) \textbf{PixelEyes}, a perception--reasoning-decoupled agent with mask-guided search, Semantic-Region BFS, and Switchable Tool Use; (2) \textbf{PixelEyes-6K}, a $5.8$K-trajectory SFT corpus distilled from a tool-augmented Gemini teacher; and (3) \textbf{Pinpoint-Bench}, a zero-hint ultra-high-resolution benchmark with diagnostic LSR/TAE metrics.
Across V*, HR-Bench, VisualProbe, MME-RealWorld-Lite, Tree-Bench, and Pinpoint-Bench, PixelEyes outperforms prior active-perception agents at both 4B and 8B scales while using fewer turns.
\section{Related Work}
\label{sec:related_work}

\noindent \textbf{Vision-Language Models and Specialized Perception.}
General-purpose VLMs -- Flamingo~\cite{flamingo}, LLaVA~\cite{llava}, GPT-4o~\cite{gpt4o}, Gemini~\cite{gemini}, the InternVL series~\cite{internvl,internvl25,internvl3,internvl35}, and the Qwen-VL series~\cite{qwenvl,qwen2vl,qwen25vl} -- align visual and textual representations and achieve strong reasoning across multimodal tasks.
Their grounding output, however, is typically a coarse bounding box and degrades on small or cluttered targets.
A parallel line of work equips MLLMs with dedicated perception modules~\cite{sam,mask2former,glamm,sa2va,hyperseg,xsam,himtok,wang2025grasp} for referring segmentation; SAMTok~\cite{samtok} unifies mask generation within the language-model interface.
These specialists ground well but tend to lose general-VQA capability when jointly fine-tuned with reasoning data -- a perception--reasoning trade-off that motivates us to treat pixel-level perception as a pluggable tool rather than a joint training target.

\noindent \textbf{Multi-turn Visual Reasoning Agents.}
Recent active-perception methods~\cite{vstar, IVM, shen2025zoomeye,Treebench,meng2025openo3} let an agent iteratively crop, zoom, or re-observe.
Early work used heuristic or tree-based search; more recent agents train this behavior with reinforcement learning~\cite{DeepEyes, Pixel-Reasoner, Adaptive-CoF, Thyme, Skywork-R1V4}.
Mini-o3~\cite{mini_o3} pushes this strategy hardest, scaling interaction turns aggressively and executing dozens of bbox crops per trajectory; ZwZ~\cite{zwz} takes the opposite route and distills zooming into a single forward pass.
Across recipes, these methods share two properties that PixelEyes drops: they rely on the base VLM's native grounding ability, and they search over coarse rectangular crops.
Both lengthen trajectories and accumulate noisy crops in the context.
Instead, we delegate grounding to a specialized perception model via a tool-call interface inspired by visual programming~\cite{gupta2023visual}, so that each module operates at its native granularity.
\section{Method: PixelEyes}
\label{sec:method}
\subsection{Problem Formulation and Agentic Pipeline}
We formulate visual evidence seeking as a multi-turn sequential decision-making process. Given a high-resolution image $I_0 \in \mathbb{R}^{H \times W \times 3}$ and a question $Q$, an autonomous agent parameterized by a VLM policy $\pi_\theta$ engages in an iterative exploratory loop to gather fine-grained evidence.

\begin{figure}[t] 
    \centering 
    \includegraphics[width=1\textwidth]{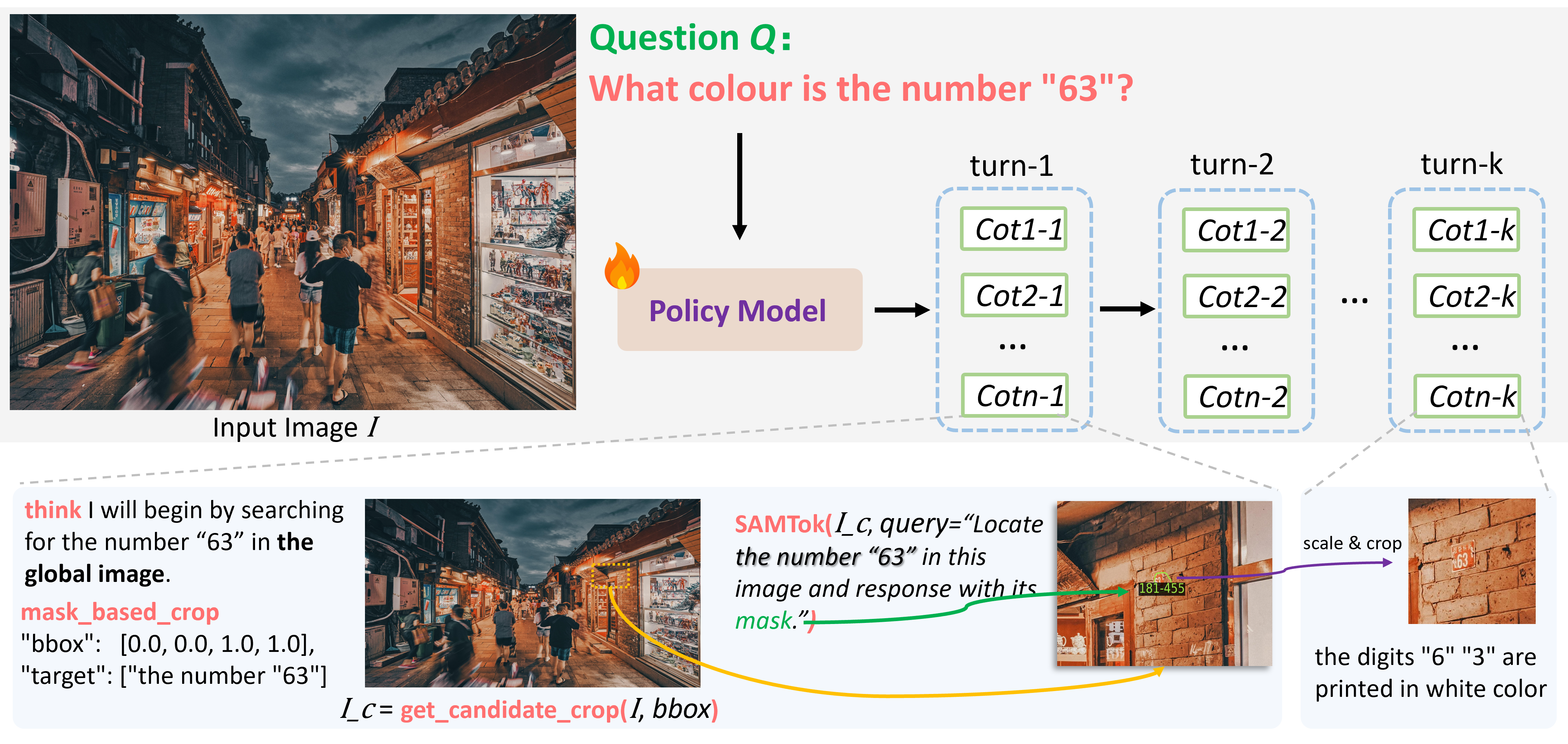}  
    \caption{\textbf{The PixelEyes Pipeline}. Given a query $Q$ and an input image $I$, the policy model generates multiple chains-of-thought per turn and invokes a mask-based crop tool with a referring expression. 
    This tool first utilizes the bbox proposal to extract a candidate crop $I_c$ from the original image $I$. Based on $I_c$, a specialized referring segmentation model (SAMTok~\cite{samtok} in this figure) returns a precise localization mask, and the cropped region is then fed back to the policy for the next turn. After $k$ turns, the agent successfully zooms onto the target and identifies its color as white. In this figure, the task is completed in only 2 turns, and the multiple turn-$i$ examples are for illustration purposes only.
    Note that the target "63" is barely visible in the global image, illustrating both the difficulty of Pinpoint-Bench and the necessity of mask-guided decoupled perception.}
    \label{fig:agent_pipeline}
\end{figure}

\begin{table*}[!t]
\centering
\setlength{\tabcolsep}{2pt}
\caption{\textbf{Comparison of Visual Search Benchmarks.} Pinpoint-Bench features an ultra-tiny Mean ROI Area (0.07\%) and is the only high-resolution benchmark providing both mask and bbox annotations for fine-grained failure analysis.}
\label{tab:dataset_comparison}
\footnotesize
\begin{tabular}{l c c c c c c c}
\toprule
Benchmark & Year & \# Samples & Image Size & Mask Annotations & BBox Annotations & Level & Mean ROI Area  \\
\midrule
V*~\cite{vstar} & 2024 & 191  & 2246×1583 & $\times$ & $\times$ & Easy & - \\
HR-Bench-4K~\cite{HR_bench} & 2025 & 800 & 4023×3503 & $\times$ & $\times$ & Easy & - \\
HR-Bench-8K~\cite{HR_bench}  & 2025 & 800 & 5727×4430 & $\times$ & $\times$ & Easy & - \\
TreeBench~\cite{Treebench} & 2026 & 450 & 2152×1615    & \checkmark& \checkmark & Hard & 3.05\% \\
VisualProbe~\cite{mini_o3} & 2026 & 515 & 5588×3676  & $\times$ & $\times$ & Hard & - \\
\midrule
\rowcolor{cyan!5}
\textbf{Pinpoint-Bench} & 2026 & 433  & 5500×3516 & \checkmark & \checkmark & Very Hard & 0.07\% \\
\bottomrule
\end{tabular}

\end{table*}

At step $t \in \{1, 2, \dots, T\}$, conditioned on the accumulated context $C_t$ (initialized as $C_1 = \{I_0, Q\}$), the model autoregressively generates a textual thought $\mathbf{H}_t$ for reasoning and planning, followed by an action $\mathbf{A}_t$:
\begin{equation}
(\mathbf{H}_t, \mathbf{A}_t) \sim \pi_\theta(\cdot \mid \mathbf{C}_t)
\end{equation}
To accommodate diverse scenarios, $\mathbf{A}_t$ utilizes a switchable tool mechanism, choosing from three operations: (1) \textbf{Mask-Based Crop}, which proposes a coarse BBox on $I_0$ alongside a referring expression for precise mask grounding and then cropping; (2) \textbf{BBox-Based Crop}, which outputs a BBox directly on $I_0$ when instance-level segmentation is inapplicable (e.g., charts or maps); and (3) \textbf{Answer}, which emits the final response to terminate the search.

Executing a cropping action returns a local image observation $\mathbf{O}_t$. This patch is appended to the context, yielding a text-and-pixel interleaved trajectory for the next turn:
\begin{equation}
\mathbf{C}_{t+1} = \mathbf{C}_t \cup \{\mathbf{H}_t, \mathbf{A}_t, \mathbf{O}_t\}
\end{equation}

\subsection{Mask-Guided Visual Search}
Due to the inherent misalignment between the general reasoning capabilities of foundational VLMs and the fine-grained localization required for tiny-instance-anchored tasks, we introduce a tool-calling mechanism to augment the base model's perception. Specifically, we employ a state-of-the-art referring segmentation model, SAMTok~\cite{samtok}, as an auxiliary visual evidence seeker. Compared to standard grounding models, SAMTok exhibits exceptional zero-shot grounding capabilities, reliably capturing precise masks even for extremely minute or irregular objects.

At step $t$, if the base model opts for a Mask-based crop, it outputs a coarse candidate bounding box $\mathbf{B}^{\text{in}}_t$ and a natural language referring expression $\mathcal{E}_t$ describing the target. The system first crops the global image $I_0$ based on $\mathbf{B}^{\text{in}}_t$ and feeds this local patch along with $\mathcal{E}_t$ into \text{SAMTok} to predict a binary mask $\mathbf{M}_t$:
\begin{equation}
\mathbf{M}_t = \text{SAMTok}(I_0[\mathbf{B}^{\text{in}}_t], \mathcal{E}_t)
\end{equation}
If \text{SAMTok} successfully grounds a target ($\mathbf{M}_t \neq \emptyset$), we compute the tight bounding box of the mask, denoted by $\mathbf{B}^{\text{mask}}_t$.
To preserve context, we enlarge the box by scaling its width and height 
around the box center with a factor of $(1+\alpha)$, yielding the target-centric crop box 
$\mathbf{B}^{\text{out}}_t$:
\begin{equation}
\mathbf{B}^{\text{out}}_t = \text{Scale}(\mathbf{B}^{\text{mask}}_t, 1 + \alpha)
\end{equation}
Conversely, if the target is absent or \text{SAMTok} fails to ground it ($\mathbf{M}_t = \emptyset$), the system triggers a fallback mechanism, directly setting $\mathbf{B}^{\text{out}}_t = \mathbf{B}^{\text{in}}_t$. The final cropped observation $\mathbf{O}_t = I_0[\mathbf{B}^{\text{out}}_t]$ is then returned to the base model for the next reasoning step. This mechanism effectively filters out background noise, ensuring highly concentrated visual inputs.

Additionally, we recognize that mask-guided visual search is not universally optimal and therefore enable switchable tool use. 
We retain the native BBox-based crop to handle regions lacking distinct instance-level semantics, where referring expressions $\mathcal{E}_t$ are difficult to formulate (e.g., charts or maps). During the agentic loop, the model autonomously evaluates the query type: it prioritizes the Mask-based crop for instance-specific pinpointing (e.g., "the color of the helmet") while reserving the BBox-based crop for structural or text-rich parsing. This hybrid approach ensures a synergetic balance between extreme local precision and general structural robustness.

\subsection{Semantic-Region BFS}
In tool-augmented multi-turn visual search tasks, conventional hierarchical search (i.e., Depth-First Search, DFS) in coupled design frameworks often forces the model into a \textit{Perception-Reflection Paradox}: it requires the base policy $\pi_\theta$ to realize it has fallen into a loop and correctly select a previous historical observation $\mathbf{O}_{k}$ ($k < t$) as the "source" for backtracking. However, if the model possessed such strong fine-grained perception and spatial reasoning to successfully reflect, it should not have missed the target in the first place. Moreover, recursive cropping creates multiple local coordinate systems across varying $\mathbf{O}_{k}$, overwhelming the model's spatial reasoning capacity. 
Consequently, the agent frequently suffers from "Inattentional Blindness" and reflection failures, trapping it in repeated cropping loops that ultimately exceed the maximum turn limit (as shown in Fig.~\ref{fig:teaser}b).

To alleviate this cognitive burden and based on the decoupled design of general reasoning and fine-grained perception, we adopt a Semantic-Region Breadth-First Search (BFS) strategy. This is built upon a strong prior: \textit{if the specialized referring perception tool (e.g., SAMTok) fails to ground the target within a proposed region (i.e., $\mathbf{M}_t = \emptyset$), the target is highly likely absent}. Consequently, the agent should immediately shift its focus to other unexamined areas rather than performing redundant local refinements.

Formally, instead of maintaining a complex tree of cropped observations, our agent tracks a flat semantic history $\mathcal{S}_t$ intrinsically embedded within the current context trajectory $\mathbf{C}_t$:
\begin{equation}
    \mathcal{S}_t = \{ (\mathbf{B}^{\text{in}}_i, c_i) \}_{i=1}^{t-1} \subset \mathbf{C}_t
\end{equation}
where $\mathbf{B}^{\text{in}}_i$ represents the historical bounding box coordinates normalized \textit{exclusively to the original image $I_0$}. Crucially, $c_i$ is a brief region caption (e.g., "the yellow SAMUDERA container" in Fig.~\ref{fig:teaser}b) generated organically during the model's textual thought process $\mathbf{H}_i$. It acts as a semantic footprint for global trajectory planning and is distinct from the referring expression $\mathcal{E}_i$, which is purely an extracted functional argument passed to the perception tool for mask grounding.

During the exploration, the policy $\pi_\theta$ entirely bypasses the "source" selection. Instead, the autoregressive generation at step $t$ follows a strict \textit{Semantic Planning $\rightarrow$ BBox Proposal $\rightarrow$ Tool Invocation} sequence. The model first assesses the semantic history $\mathcal{S}_t$ and articulates a new semantically plausible region via caption $c_t$ within its thought $\mathbf{H}_t$. Immediately after this conceptual planning, it generates the specific global coordinates $\mathbf{B}^{\text{in}}_t$ ($\subset I_0$) and constructs the tool-calling action $\mathbf{A}_t$ (e.g., \texttt{mask\_based\_crop}), taking $\mathbf{B}^{\text{in}}_t$ and a precise referring expression $\mathcal{E}_t$ as parameters:
\begin{equation}
    c_t \in \mathbf{H}_t, \quad \mathbf{A}_t(\mathbf{B}^{\text{in}}_t, \mathcal{E}_t) \sim \pi_{\theta}(\cdot \mid \mathbf{C}_t)
\end{equation}

If $\mathbf{M}_t = \emptyset$, this sequence (including the planning caption $c_t$) natively becomes part of the updated context $\mathbf{C}_{t+1}$. Guided by the transparent history of explored semantic regions ($c_{i<t}$) and failed coordinates ($\mathbf{B}^{\text{in}}_{i<t}$), the agent inherently conducts a BFS-style horizontal exploration to a fresh area in the subsequent turn. By utilizing thoughts to plan spatial coverage and anchoring all coordinates to a single global reference frame ($I_0$), Semantic-Region BFS effectively prevents the agent from getting lost in recursive local crops and ensures fluent, coverage-prioritized visual exploration. A depth-first verification can be naturally performed when the agent encounters strong visual cues.

\subsection{PixelEyes-6K Dataset and Model Training}
To validate our PixelEyes framework and ensure a fair and rigorous comparison, we also adopt a 2-stage training strategy and strictly align our training data sources with Mini-o3~\cite{mini_o3}, introducing no additional data. Instead, we use our novel data pipeline—incorporating \textit{mask-guided visual search}, \textit{switchable tool use}, and \textit{semantic-region BFS logic}—to resynthesize trajectories and ultimately produce the \textbf{PixelEyes-6K} dataset.

\begin{figure}[t]
    \vspace{-3mm}
    \centering 
    \includegraphics[width=1.00\textwidth]{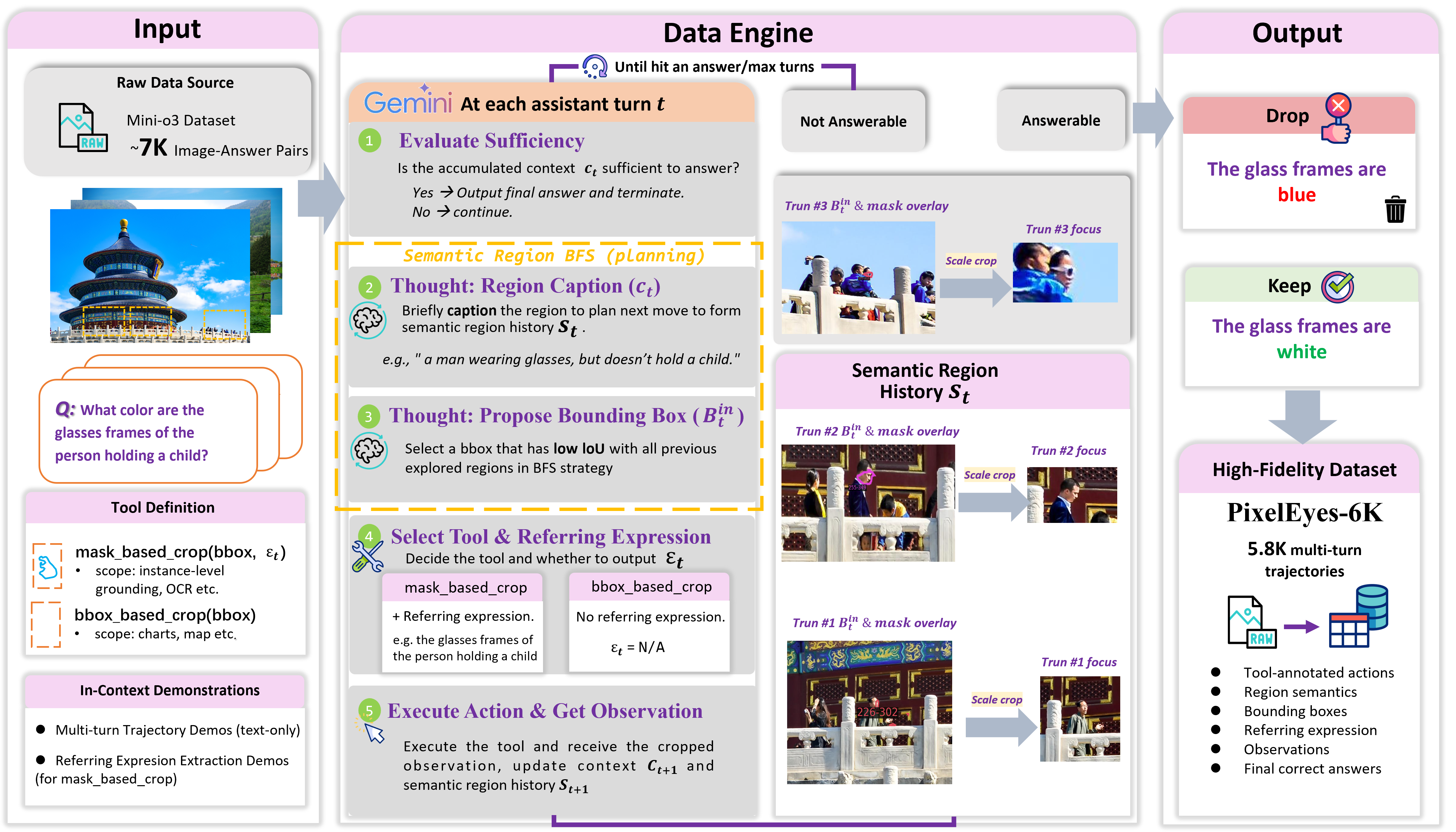}
    \caption{Data Pipeline for PixelEyes-6K Dataset Construction. (1) \textbf{In-Context Initialization:} Manually annotated multi-turn trajectories and specific referring expression examples are provided to guide Gemini's reasoning. (2) \textbf{Interactive Closed-Loop Rollout:} Incorporating our \textit{mask-guided visual search} and \textit{Semantic-Region BFS} strategies, the VLM iteratively plans its next move, generates region captions ($c_t$), proposes low-IoU bounding boxes ($\mathbf{B}^{\text{in}}_t$), and dynamically invokes tools. (3) \textbf{Trajectory Filtering:} Only trajectories resulting in the correct answer are retained, distilling 5.8K high-fidelity trajectories from 7K raw pairs.}
    \label{fig:data_pipe}
\end{figure}

\textbf{Synthesize Expert Trajectories.} To empower our base model with fine-grained perception and high-efficiency visual exploration, we leverage Gemini's reasoning capabilities as the base VLM for global planning, augmented by SAMTok's precise localization. This directly instantiates our \textit{mask-guided visual search} paradigm in data synthesis.

As illustrated in Fig.~\ref{fig:agent_pipeline}, the expert trajectories are synthesized via an interactive, closed-loop rollout. Before Gemini's actual inference, we formally define the \texttt{mask\_based\_crop} and \texttt{bbox\_based\_crop} tools along with their respective applicability scopes.
To guide Gemini's reasoning logic and ensure structured outputs, we manually annotate several text-only multi-turn trajectories as in-context demonstrations. Additionally, to teach the model how to precisely extract the referring expression ($\mathcal{E}_t$) from the question for the \texttt{mask\_based\_crop} tool, we provide dedicated in-context examples (while \texttt{bbox\_based\_crop} requires no target parameter). 

During the rollout, at each assistant turn $t$, Gemini is prompted to adopt an action after evaluating whether the accumulated context $\mathbf{C}_t$ is sufficient to answer the question and terminate. If the visual evidence is insufficient, following the \textit{Semantic-Region BFS} strategy, Gemini must first generate a brief region caption ($c_t$) within its thought to explicitly plan its next move. This is immediately followed by proposing a specific global bounding box ($\mathbf{B}^{\text{in}}_t$) that maintains a low Intersection over Union (IoU) with all previously explored regions in the semantic region history $\mathcal{S}_t$ (excluding the initial full-image grounding box). Concurrently, the model evaluates the task type to autonomously select the appropriate tool (e.g., \texttt{mask\_based\_crop} vs. \texttt{bbox\_based\_crop}) and decides whether to output a referring expression to attempt precise target localization. This entire sequence is dynamically appended to the context, incrementally expanding the semantic region history $\mathcal{S}_t$ and inherently driving a BFS-style horizontal exploration to unexamined areas, thereby constructing realistic and context-coherent multi-turn interactive trajectories.

Finally, to filter out potential hallucinations or erroneous reasoning steps, we strictly retain only those trajectories where Gemini successfully produces the correct final answer. Applying this rigorous closed-loop pipeline to approximately 7K raw image-question pairs from the Mini-o3 dataset, we distill 5.8K high-fidelity, context-coherent multi-turn trajectories, forming our \textbf{PixelEyes-6K} dataset.

\textbf{Training Stage 1: Supervised Fine-Tuning (SFT).} We fine-tune the base model on our PixelEyes-6K dataset, enabling it to master diverse search strategies, tool-switching patterns, and semantic region BFS search logic, thus laying a solid foundation for subsequent reinforcement learning.

\textbf{Training Stage 2: Reinforcement Learning (RL).}
Following Mini-o3, we use Group Relative Policy Optimization (GRPO)~\cite{GRPO}.
To improve efficiency, we filter the raw RL dataset with our SFT model, removing outliers that are either unsolved (solve rate $=0$) or trivial (solve rate $>0.9$), leaving 5.5K samples.
Unlike Mini-o3, which uses over-turn masking to encourage trial-and-error, our framework prioritizes efficient search.
We therefore discard over-turn masking and use vanilla GRPO to optimize concise visual evidence-seeking paths.

\subsection{Pinpoint-Bench: A Zero-Hint Evaluation Frontier}
\label{sec:benchmark}
Recent visual search benchmarks (e.g., V* ~\cite{vstar}, HR-Bench~\cite{HR_bench}) are beginning to show signs of saturation, with some models already exceeding 90\% accuracy. Conversely, other evaluations, such as TreeBench~\cite{Treebench} and MME-RealWorld~\cite{MME}, emphasize logical reasoning rather than pure, extreme-scale visual search, often featuring targets that are neither sufficiently minute nor embedded in ultra-high-resolution contexts. While the VisualProbe ~\cite{mini_o3} dataset offers appropriate difficulty, its complete lack of spatial annotations (e.g., bounding boxes or masks) creates immense friction for researchers—requiring manual re-discovery of "needle-in-a-haystack" targets simply to analyze failure cases. Furthermore, in such extreme scenarios, minor linguistic ambiguities in the ground-truth answers frequently lead to unjust penalization by LLM judges, obscuring a model's true localization.

To address these critical gaps and rigorously evaluate the limits of active visual search, we introduce \textbf{Pinpoint-Bench}, a meticulously human-annotated benchmark comprising 433 high-resolution samples across diverse, heavily cluttered scenes. Detailed statistics are presented in Appendix Fig. \ref{fig:statistics}. Our benchmark pushes the evaluation frontier through the following defining characteristics:

\noindent \textbf{Extreme Complexity and Zero-Hint Protocol.}
The benchmark is constructed exclusively from ultra-high-resolution images ($\ge 4\text{k}$) cluttered with severe distractors. The target objects are extremely minute, typically occupying a minuscule fraction ($\sim0.07\%$ as presented in Tab.~\ref{tab:dataset_comparison}) of the image. Crucially, we enforce a strict \textbf{zero-hint} protocol: all queries are entirely devoid of spatial priors (e.g., "in the bottom-left corner") or salient macro-anchors (e.g., "on the huge bridge"). Covering single-object attribute recognition, minute OCR, and spatial relationships, these tasks force models to rely entirely on autonomous, exhaustive visual search rather than text-guided shortcuts.

\noindent \textbf{Exhaustive Annotations and Diagnostic Metrics.}
A unique contribution of Pinpoint-Bench is the inclusion of exhaustive bounding-box and instance-mask annotations for each target. This directly eliminates the prohibitive cost of manual case analysis, allowing researchers to automatically extract and evaluate intermediate reasoning trajectories. Leveraging these dense annotations, we introduce two novel diagnostic metrics. \textbf{Turn-to-Answer Efficiency (TAE)} evaluates the trade-off between performance and interaction cost by measuring accuracy normalized by the average number of interaction turns, formulated as:
\begin{equation}
\text{TAE} = \frac{\text{Accuracy}}{\text{AvgTurns}}
\label{eq:tae}
\end{equation}
A higher TAE indicates a highly efficient trajectory that reaches the correct answer with minimal reasoning steps. \textbf{Localization Success Rate (LSR)} tracks whether essential visual evidence is actually "discovered" during the exploration. Formally, let $N$ be the total number of samples, and $T_i$ be the number of bounding boxes (crops) in the trajectory of the $i$-th sample. For each sample, let $B_{i,t}$ denote the bounding box region of the $t$-th crop, and $\{M_{i,k}\}$ be the set of ground-truth target instance masks. A sample is considered a localization success if any historical bounding box overlaps with any ground-truth mask at the pixel level. LSR is thus defined as:
\begin{equation}
\text{LSR} = \frac{1}{N} \sum_{i=1}^{N} \mathbf{1} \left[ \exists t \in \{1, \dots, T_i\}, \exists k, \sum_{(x,y) \in B_{i,t}} M_{i,k}(x,y) > 0 \right]
\label{eq:lsr}
\end{equation}
where $\mathbf{1}[\cdot]$ is the indicator function. By checking if the crop area contains at least one mask pixel, LSR considers a trial successful if \textit{any} crop within the trajectory covers the target instance, regardless of the final textual answer. 

By bypassing the interference of long-context degradation, LSR specifically quantifies the model's pure "visual evidence seeking" capability. Consequently, the gap between LSR and Accuracy serves as a powerful diagnostic tool to quantify \textbf{Inattentional Blindness}—revealing whether a failure originates from a perception bottleneck (finding the target but failing to recognize or utilize the evidence) or a reasoning bottleneck (failing to discover the target through effective exploration and planning).

\noindent \textbf{Robust Multi-Alias Scoring.}
In highly challenging visual search cases, natural language descriptions of visual attributes are inherently ambiguous (e.g., a "beige" backpack might be reasonably described as "white" or "light brown"). To prevent models that successfully locate the target from being unjustly penalized by rigid textual ground truths, we construct exhaustive multi-alias answer sets for each query. Integrated with an LLM-based judge, this robust scoring mechanism effectively prevents linguistic nuances from overshadowing successful search efforts, offering a highly faithful reflection of a model's true active perception capability.

\section{Experiments}
\label{sec:exp}
\subsection{Experimental Setups}
\noindent \textbf{Benchmarks and Baselines.}
We evaluate on three groups of benchmarks. (1) \emph{Active perception}: V*~\cite{vstar} and HR-Bench-4K/8K~\cite{HR_bench} for high-resolution visual search. (2) \emph{Complex visual search}: the VisualProbe suite (Easy/Medium/Hard)~\cite{mini_o3} and our Pinpoint-Bench for needle-in-a-haystack localization. (3) \emph{General and structural reasoning}: MME-RealWorld-Lite~\cite{MME} and Tree-Bench~\cite{Treebench} to check that decoupling does not erode general multimodal capability.
We compare against a closed-source frontier model (Gemini-3-Flash~\cite{gemini}), open-source foundation VLMs (Qwen-2.5-VL and the Qwen-3-VL series~\cite{qwen25vl, qwen3}), and specialized active-perception agents (DeepEyes, Pixel-Reasoner, Thyme, Mini-o3~\cite{DeepEyes, Pixel-Reasoner, Thyme, mini_o3}).
A training-free plug-in variant of our protocol is reported in Tab.~\ref{tab:appendix_plug} (Appendix).

\noindent \textbf{Metrics.}
We report standard accuracy on all benchmarks, and additionally LSR and TAE (Sec.~\ref{sec:benchmark}) on Pinpoint-Bench to separate localization from answering.

\noindent \textbf{Training and Testing.} Hyperparameters and hardware are deferred to Sec.~\ref{sec:appendix_more_exp}.

\begin{table}[t]
\small
\centering
\caption{Quantitative comparison of different methods on visual search benchmarks. The best results are highlighted in bold. MME-R-L denotes the MME-RealWorld-Lite~\cite{MME} benchmark. TAE and LSR require localization traces and are only reported for methods that emit them; Qwen-3-VL-235B was evaluated on the V*/HR-Bench subset only.}
\label{tab:integrated_full_final}
\resizebox{\textwidth}{!}{
\begin{tabular}{llccccccccccc}
\toprule
\multirow{2}{*}{Model} & \multirow{2}{*}{Size} & \multirow{2}{*}{V*} & \multirow{2}{*}{HR-4K} & \multirow{2}{*}{HR-8K} & \multicolumn{3}{c}{VisualProbe} & \multicolumn{3}{c}{Pinpoint-Bench} & \multirow{2}{*}{MME-R-L} & \multirow{2}{*}{Tree-Bench} \\
\cmidrule(lr){6-8} \cmidrule(lr){9-11}
 & & & & & Hard & Medium & Easy & Acc. & TAE & LSR & & \\
\midrule
\multicolumn{13}{c}{Closed-source Models} \\
\midrule
Gemini-3-Flash~\cite{gemini} & - & 84.82 & 89.25 & 85.50 & 47.17 & 50.75 & 67.38 & 42.26 & - & - & 60.34 & 56.54 \\
\midrule
\multicolumn{13}{c}{Open-source Base Models} \\
\midrule
Qwen-2.5-VL~\cite{qwen25vl} & 7B & 75.50 & 68.20 & 62.70 & 23.90 & 26.00 & 39.10 & 39.03 & - & - & 44.37 & 41.48 \\
Qwen-3-VL~\cite{qwen3} & 4B & 80.10 & 78.25 & 72.88 & 34.91 & 40.30 & 56.74 & 46.19 & - & - & 44.55 & 42.71 \\
Qwen-3-VL~\cite{qwen3} & 8B & 86.39 & 78.88 & 74.63 & 51.89 & 40.67 & 65.25 & 49.88 & - & - & 49.04 & 46.91 \\
Qwen-3-VL~\cite{qwen3} & 235B & 87.96 & 84.50 & 81.62 & - & - & - & - & - & - & - & - \\
\midrule
\multicolumn{13}{c}{Expert Active Agents} \\
\midrule
Pixel-Reasoner~\cite{Pixel-Reasoner} & 7B & 86.30 & 74.00 & 66.90 & 28.80 & 29.60 & 58.40 & 29.79 & 15.56 & 46.88 & 54.32 & 40.98 \\
Thyme~\cite{Thyme} & 7B & 82.20 & 77.00 & 72.00 & 46.23 & 43.28 & 62.41 & 40.42 & - & - & 50.13 & 39.75 \\
DeepEyes~\cite{DeepEyes} & 7B & 83.30 & 73.20 & 69.50 & 35.10 & 29.80 & 60.10 & 39.72 & 14.89 & 20.79 & 53.53 & 37.28 \\
Mini-o3~\cite{mini_o3} & 7B & 85.34 & 71.75 & 67.50 & 45.28 & 48.51 & 63.12 & 44.34 & 8.38 & \textbf{78.52} & 42.26 & 40.25 \\
\midrule
\multicolumn{13}{c}{Ours} \\
\midrule
PixelEyes & 4B & 91.62 & 81.75 & 79.88 & 54.72 & 55.22 & 68.79 & 54.73 & 26.13 & 76.91 & 54.51 & 45.93 \\
$\Delta$ vs. Qwen3-VL-4B & & \up{11.52} & \up{3.50} & \up{7.00} & \up{19.81} & \up{14.92} & \up{12.05} & \up{8.54} & - & - & \up{9.96} & \up{3.22} \\
\midrule
PixelEyes & 8B & \textbf{94.24} & \textbf{85.00} & \textbf{83.15} & \textbf{59.44} &  \textbf{55.22} & \textbf{71.63} & \textbf{55.20} & \textbf{26.64} & 74.83 & \textbf{59.25} & \textbf{48.40} \\
$\Delta$ vs. Qwen3-VL-8B & & \up{7.85} & \up{6.12} & \up{8.52} & \up{7.55} & \up{14.55} & \up{6.38} & \up{5.32} & - & - & \up{10.21} & \up{1.49} \\

\bottomrule
\end{tabular}
}
\end{table}

\subsection{Main Results}

\noindent \textbf{Comparison with baselines.}
Tab.~\ref{tab:integrated_full_final} reports accuracy across the three benchmark groups.
At the 4B scale, PixelEyes improves over its Qwen-3-VL-4B base by $+11.5$ on V*, $+7.0$ on HR-Bench-8K, and $+19.8/+14.9/+12.0$ on VisualProbe Hard/Medium/Easy.
At the 8B scale, PixelEyes reaches $94.24\%$ on V*, $85.00\%$ on HR-Bench-4K, and $83.15\%$ on HR-Bench-8K -- higher than the much larger Qwen-3-VL-235B on the three benchmarks.
Against prior active-perception agents (Pixel-Reasoner, Thyme, DeepEyes, Mini-o3), PixelEyes-4B is higher on every accuracy column we measure, and PixelEyes-8B is best overall.
On the general/structural reasoning group, PixelEyes-8B also improves over its base by $+10.2$ on MME-RealWorld-Lite and $+1.5$ on Tree-Bench, indicating that decoupling perception from reasoning does not erode broader multimodal capability.

\noindent \textbf{Pinpoint-Bench analysis.}
The zero-hint setting separates models more sharply. Qwen-3-VL-4B and 8B reach only $46.19\%$ and $49.88\%$ accuracy. Prior agents improve localization but not always answering: For example, Mini-o3 attains $\text{LSR}=78.52\%$ -- it does find the target -- yet its accuracy is $44.34\%$ and its $\text{TAE}=8.38$. The $\text{LSR}-\text{Acc}$ gap of $34$ points quantifies inattentional blindness directly. 
PixelEyes-4B trades a small LSR drop ($-1.61$) for $+10.39$ Acc. and $+17.75$ TAE, and PixelEyes-8B pushes Acc and TAE further to $55.20$ and $26.64$. The takeaway is that pixel-tight mask crops not only find evidence -- they also make the evidence usable to the reasoner.

\subsection{Ablation Study}
We conduct ablation studies on our SFT data, RL training, switchable tool use, search strategy, and mask grounder. Benchmarking is primarily evaluated on HR-Bench 4K/8K, VisualProbe and our constructed Pinpoint-Bench, utilizing Qwen-3-VL-4B as the default base model. Unless explicitly stated otherwise, all reported results correspond to the performance after RL.

\noindent \textbf{SFT data.}
Tab. \ref{tab:sft} contrasts fine-tuning Qwen-3-VL on Mini-o3 versus PixelEyes-6K. Mini-o3's data drops VisualProbe-Hard from 34.91 to 24.52 ($-10.4$). We observe that Mini-o3 induces a rigid DFS behavior: the model chronically trapped itself in redundant loops—repeatedly cropping identical or nested sub-regions—until exhausting the maximum allowed turns. Conversely, fine-tuning on PixelEyes-6K reaches 50.94 ($+16.0$). This confirms that the structural quality of trajectories, rather than volume, is the key lever for visual reasoning.

\noindent \textbf{RL.}
Adding vanilla GRPO on top of our SFT model lifts VisualProbe-Hard to $54.72$ and Pinpoint-Bench Acc. to $54.73$ (Tab.~\ref{tab:sft}).

\noindent \textbf{Search strategy.}
To evaluate the performance gains of our Mask-Guided Visual Search and Semantic-Region BFS, we conduct an ablation study in Tab.~\ref{tab:strategy} (SFT only). In the "A+B" formulation, A denotes the cropping strategy (bbox or mask) and B indicates the candidate proposal strategy. Specifically, "BFS" represents our Semantic-Region BFS; "Free" relaxes the low-overlap constraint for trajectory proposals, searching the entire image without a reference observation; and "DFS" strictly couples each proposal with its source observation. Note that "DFS" is inapplicable to mask-based cropping, as it outputs local crops of potential targets rather than observation-forming bboxes, eliminating the need for subsequent verification. Importantly, BFS and DFS here do not denote the standard graph algorithms, but rather signify whether explicit zoom-in verification is enforced.

Overall, as shown in Tab.~\ref{tab:strategy}, the Mask-Guided Visual Search strategy consistently outperforms the pure bbox-based search strategies. Notably, under the bbox-based setting, there is no significant performance gap among BFS, DFS, and Free. We attribute this primarily to the limitation that under a pure bbox zoom-in strategy, the model cannot confidently rely on its previous perceptual results to explicitly exclude already searched regions. In contrast, under the mask mode, since perception is entirely delegated to the mask grounder—whose perceptual capability surpasses that of the base model itself—the BFS strategy can yield noticeable performance improvements.

\noindent \textbf{Switchable tool use.}
On HR-Bench, where many images are charts or documents, the mask-only variant of PixelEyes already improves over the Qwen-3-VL base (Tab.~\ref{tab:swi}). 
Enabling the bbox fallback adds $+1.75$ on HR-Bench-4K and $+0.38$ on HR-Bench-8K, confirming that allowing the policy to revert to bbox cropping for structure-heavy queries is beneficial.

\noindent \textbf{Grounding backends.}
We further ablate the referring segmentation backend in Tab.~\ref{tab:backends}. 
Replacing SAMTok with Sa2VA consistently degrades Acc., TAE, and LSR, indicating that PixelEyes depends critically on robust mask grounding. 
Specifically, Sa2VA struggles to localize small target regions in high-resolution scenarios, often failing to return valid masks and forcing premature fallback to bounding-box cropping. 
In contrast, SAMTok achieves a robust tool Invoke Success Rate (ISR) of 99.17\% (476/480 calls), confirming it is not merely an interchangeable implementation choice but a key component enabling precise high-resolution visual grounding.

\begin{table}[!ht]
\centering
\small
\setlength{\tabcolsep}{3pt}

\begin{minipage}[b]{0.52\textwidth}
    \centering
    \caption{Ablation on SFT data and RL training.}
    \label{tab:sft}
    \resizebox{0.9\textwidth}{!}{
        \begin{tabular}{lcccc}
        \toprule
        \multirow{2}{*}{Model} & \multicolumn{3}{c}{VisualProbe} & \multirow{2}{*}{Pinpoint} \\
        \cmidrule(lr){2-4}
         & Hard & Medium & Easy &   \\
        \midrule
        Qwen-3-VL~\cite{qwen3} & 34.91 & 40.30 & 56.74 & 46.19  \\
        \midrule
        w/ Mini-o3 SFT & 24.52 & 33.58 & 38.29 & 29.56  \\
        w/ Our SFT & 50.94 & 52.24 & 68.09 & 52.66  \\
        w/ Our SFT+RL & 54.72 & 55.22 & 68.79 & 54.73    \\
        \bottomrule
        \end{tabular}
    }
\end{minipage}
\hfill
\begin{minipage}[b]{0.43\textwidth}
    \centering
    \caption{Ablation on search strategies.}
    \label{tab:strategy}
    \resizebox{0.77\textwidth}{!}{
        \begin{tabular}{lccc}
        \toprule
        Model & Acc. & TAE & LSR    \\
        \midrule
        Mask+BFS  & 52.66 & 25.31 & 75.98  \\
        \midrule
        Mask+Free  & 50.58 & 24.33 & 66.74  \\
        BBox+BFS  & 48.73 & 20.43 & 68.13  \\
        BBox+DFS  & 48.97 & 23.56 & 65.13  \\
        BBox+Free  & 48.51 & 21.54 & 68.59  \\
        \bottomrule
        \end{tabular}
    }
\end{minipage}
\end{table}

\begin{table}[!ht]
\centering
\small
\setlength{\tabcolsep}{3pt}
\begin{minipage}[b]{0.52\textwidth}
    \centering
    \caption{Ablation on switchable tool use.}
    \label{tab:swi}
    \resizebox{0.6\textwidth}{!}{
        \begin{tabular}{lcc}
        \toprule
        Model & HR-4K & HR-8K \\
        \midrule
        Qwen-3-VL~\cite{qwen3} & 78.25 & 72.88 \\
        \midrule
        w/o Switchable & 80.00 & 79.50 \\
        w/ Switchable & 81.75 & 79.88 \\
        \bottomrule
        \end{tabular}
    }
\end{minipage}
\hfill
\begin{minipage}[b]{0.43\textwidth}
    \centering
    \caption{Ablation on grounding backends.}
    \label{tab:backends}
    \resizebox{0.8\textwidth}{!}{
        \begin{tabular}{lcccc}
        \toprule
        Model & Acc. & TAE & LSR & ISR \\
        \midrule
        SAMTok~\cite{samtok} & 54.73 & 26.13 & 76.91 & 99.17 \\
        Sa2VA~\cite{sa2va} & 46.19 & 21.41 & 50.58 & 65.29 \\
        \bottomrule
        \end{tabular}
    }
\end{minipage}
\end{table}

\section{Conclusion}
\label{sec:conclusion}
This paper introduces \textbf{PixelEyes}, an active visual reasoning agent built on a perception-reasoning decoupling paradigm.
By delegating fine-grained localization to SAMTok and preserving high-level reasoning within a strong general-purpose VLM, PixelEyes alleviates the limitations of coupled visual-search systems, including inefficient exploration and inattentional blindness.
We further introduce a mask-guided visual search mechanism enhanced by \textit{Semantic-Region BFS} and \textit{Switchable Tool Use}, enabling precise, efficient, and broadly applicable evidence acquisition across diverse visual scenarios.
To train models under this paradigm, we develop a high-fidelity trajectory synthesis engine that augments Gemini-3-Flash with the \texttt{mask\_based\_crop} perception tool, producing the \textbf{PixelEyes-6K} dataset.
We also present \textbf{Pinpoint-Bench}, a challenging ultra-high-resolution benchmark under a strict ``zero-hint'' protocol, with dense mask and bounding-box annotations for diagnostic evaluation through \textit{Location Success Rate} (LSR) and \textit{Turn-To-Answer Efficiency} (TAE).
Extensive experiments demonstrate that PixelEyes achieves state-of-the-art accuracy and efficiency, suggesting that explicitly optimizing the structure and logic of visual evidence seeking is a promising path toward more reliable active visual reasoning agents.

{\small
\bibliographystyle{ieee_fullname}
\bibliography{ref_bib}
}

\appendix
\newpage
\appendix

\begin{center}
    {\LARGE \textbf{Appendix}}
\end{center}

\section{More Experiment Results}
\label{sec:appendix_more_exp}

\noindent \textbf{Implementation Details.} 
For SFT, we fine-tune the base model for 1 epoch using the AdamW optimizer with a learning rate of 2e-5, weight decay of 0.05, and $(\beta_1, \beta_2) = (0.9, 0.999) $. We use a warmup ratio of 0.03, gradient clipping with a maximum norm of 1.0, and a batch size of 32.
For RL, we adopt the GRPO algorithm with a learning rate of 1e-6 and a group size of 16. The policy is optimized using the seq-mean-token-mean loss. To stabilize training, we apply clipping thresholds of 0.2 and 0.3 for the lower and upper bounds, respectively. 
Training is performed with a global batch size of 64, a mini-batch size of 32, and a per-device micro-batch size of 1. We do not employ KL divergence or entropy regularization. The visual input resolution ranges from 40K to 2M pixels. 
To balance computational cost and long-context capability, we cap each dialogue at 6 turns for training and set the maximum context length to 10,240 tokens. 
We use Qwen3-VL-8B as the reward model during RL training, while Gemini-3-Flash is adopted as the LLM judge for benchmarking. 
%

%
For testing, we conduct experiments on 4 NVIDIA H100 GPUs, with a maximum of 6 interaction rounds per query. All baseline methods are evaluated using their officially released implementations without any modifications.

\begin{table}[h]
\small
\centering
\caption{The training-free results demonstrate significant performance improvements by integrating our tools in a plug-and-play manner into Gemini’s multi-turn reasoning process. MME-R-L denotes the MME-RealWorld-Lite~\cite{MME} benchmark.}
\label{tab:appendix_plug}
\resizebox{\textwidth}{!}{
\begin{tabular}{llccccccccccc}
\toprule
\multirow{2}{*}{Model} & \multirow{2}{*}{Size} & \multirow{2}{*}{V*} & \multirow{2}{*}{HR-4K} & \multirow{2}{*}{HR-8K} & \multicolumn{3}{c}{VisualProbe} & \multicolumn{3}{c}{Pinpoint-Bench} & \multirow{2}{*}{MME-R-L} & \multirow{2}{*}{Tree-Bench} \\
\cmidrule(lr){6-8} \cmidrule(lr){9-11}
 & & & & & Hard & Medium & Easy & Acc. & TAE & LSR & & \\
\midrule
\multicolumn{13}{c}{Closed-source Models} \\
\midrule
Gemini-3-Flash~\cite{gemini} & - & 84.82 & 89.25 & 85.50 & 47.17 & 50.75 & 67.38 & 42.26 & - & - & 60.34 & 56.54 \\
\midrule
Gemini-3-Flash Tool & - & 97.37 & 88.63 & 91.12 & 61.32 & 68.28 & 77.30 & 68.36 & 44.95 & 89.97 & 63.42 & 60.00 \\
$\Delta$ \textit{v.s.} Gemini-3-Flash & & \up{12.55} & \down{0.62} & \up{5.62} & \up{14.15} & \up{17.53} & \up{9.92} & \up{26.10} & - & - & \up{3.08} & \up{3.46} \\
\bottomrule
\end{tabular}
}
\end{table}

\noindent \textbf{Plug-and-play Deployment Demonstrates Strong Transferability.}
The transferability of our framework is further validated through the \textit{Gemini-3-Flash Tool} variant in Tab.~\ref{tab:appendix_plug}. 
Specifically, we integrate our training-free, mask-guided search mechanism directly into the Gemini API as a plug-and-play component, without any additional fine-tuning or parameter updates. 
As the results show, this simple augmentation consistently yields substantial performance gains across all benchmarks. 
For instance, we observe improvements of +14.15\% on VisualProbe-Hard and +26.10\% on Pinpoint-Bench Acc., along with consistent gains on other evaluation settings. 
These results demonstrate that our framework is highly transferable, model-agnostic, and can effectively enhance strong proprietary VLMs through inference-time augmentation alone.

\section{Deeper analysis on Pinpoint-Bench}
\noindent \textbf{Explicit Failure Decomposition Quantifies Inattentional Blindness.} 
To pinpoint the exact failure modes, we decompose the results on Pinpoint-Bench into three mutually exclusive categories in Tab.~\ref{tab:failure_decomposition}. 
Crucially, while Mini-o3 achieves a high LSR ($78.52\%$), it tends to output overly coarse, large bounding boxes, leading to a massive $38.34\%$ inattentional blindness rate. 
In contrast, our first-round global grounding utilizes the precise SAMTok mask-guided results to compute tight, accurate bounding boxes, resulting in a higher localized-and-correct rate ($50.35\%$) and a more reasonable LSR-Acc gap ($22.17\%$). 
These results mathematically prove that a substantial fraction of VLM errors stem from post-localization reasoning failures, and highlight the advantage of our fine-grained localization mechanism.

\noindent \textbf{Task-Specific Analysis Highlights Granular Mask Benefits.} 
To further understand where our decoupled mechanism excels, we break down the performance across three distinct task types in Tab.~\ref{tab:task_breakdown}: Attribute Recognition, Spatial Relation, and OCR. 
Empirical results show that PixelEyes-4B-RL achieves substantial advantages in Attribute Recognition ($56.52\%$) and OCR ($54.62\%$), outperforming Mini-o3 by $+12.25\%$ and $+11.54\%$, and Qwen-3-VL-4B by $+12.65\%$ and $+9.24\%$, respectively. 
This pronounced gain demonstrates that our training-free, mask-guided search mechanism yields the maximum dividend in scenarios requiring fine-grained local visual features and precise text-region localization. 
Conversely, in Spatial Relation tasks, Qwen-3-VL-4B maintains an edge ($60.00\%$), suggesting that resolving complex relative positions still heavily relies on global layout understanding. 

\noindent \textbf{Turn-Level Efficiency Analysis and Headline Figures.} 
To make our efficiency arguments tangible and unpack the composite TAE (Turn-Aware Efficiency) metric, we explicitly report the primitive turn statistics on Pinpoint-Bench in Tab.~\ref{tab:efficiency_turns}. 
The empirical results deliver a striking ``headline number'' regarding operational efficiency: PixelEyes-4B-RL requires only \textbf{2.09 average turns} per sample, which is dramatically fewer than Mini-o3's \textbf{5.29 average turns}, while simultaneously achieving a $+10.39\%$ absolute gain in Accuracy ($54.73\%$ vs. $44.34\%$). 
Furthermore, when compared to DeepEyes, our framework not only demonstrates superior accuracy ($54.73\%$ vs. $39.72\%$) but also accomplishes this with a shorter interaction trajectory ($2.09$ vs. $2.67$ average turns). 
Although PixelReasoner exhibits a slightly lower average turn count ($1.91$), its reasoning path yields a severely degraded accuracy of only $29.79\%$. 
These clear-cut comparisons establish that PixelEyes-4B-RL hits a sweet spot of high accuracy and low communication overhead, reinforcing that our mask-guided mechanism significantly streamlines the multi-turn visual grounding process.

\begin{table}[t]
\centering
\caption{Performance comparison on localization and QA.}
\label{tab:failure_decomposition}
\resizebox{\textwidth}{!}{
\begin{tabular}{lccc}
\toprule
\textbf{Method} & \textbf{Loc. Success \& Correct} & \textbf{Loc. Success \& Incorrect} & \textbf{Not Localized}  \\
\midrule
PixelReasoner~\cite{Pixel-Reasoner}   & 71 / 433 (16.40\%)  & 132 / 433 (30.48\%) & 230 / 433 (53.12\%) \\
DeepEyes~\cite{DeepEyes}        & 44 / 433 (10.16\%)  & 46 / 433 (10.62\%)  & 343 / 433 (79.21\%)  \\
Mini-o3~\cite{mini_o3}         & 174 / 433 (40.18\%) & 166 / 433 (38.34\%) & 93 / 433 (21.48\%) \\
PixelEyes-4B-RL & 218 / 433 (50.35\%) & 115 / 433 (26.56\%) & 100 / 433 (23.09\%)  \\
\bottomrule
\end{tabular}%
}
\end{table}

\begin{table}[t]
\centering
\caption{Performance breakdowns across different task types on Pinpoint-Bench.}
\label{tab:task_breakdown}
\resizebox{\textwidth}{!}{
\begin{tabular}{lcccc}
\toprule
\textbf{Method} & \textbf{Attribute (n=253)} & \textbf{Spatial Relation (n=50)} & \textbf{OCR (n=130)} & \textbf{Overall (n=433)} \\
\midrule
Qwen-3-VL-4B~\cite{qwen3}    & 111 / 253 (43.87\%) & 30 / 50 (60.00\%) & 59 / 130 (45.38\%) & 200 / 433 (46.19\%) \\
Mini-o3~\cite{mini_o3}         & 112 / 253 (44.27\%) & 24 / 50 (48.00\%) & 56 / 130 (43.08\%) & 192 / 433 (44.34\%) \\
PixelEyes-4B-RL & 143 / 253 (56.52\%) & 23 / 50 (46.00\%) & 71 / 130 (54.62\%) & 237 / 433 (54.73\%) \\
\bottomrule
\end{tabular}%
}
\end{table}

\begin{table}[t]
\centering
\caption{Efficiency and turn-level performance analysis on Pinpoint-Bench}
\label{tab:efficiency_turns}
\resizebox{0.65\textwidth}{!}{
\begin{tabular}{lcccc}
\toprule
\textbf{Method} & \textbf{Acc.} & \textbf{Total Turns} & \textbf{Avg. Turns} & \textbf{TAE} \\
\midrule
PixelReasoner~\cite{Pixel-Reasoner}   & 29.79 & 829  & 1.91 & 15.56 \\
DeepEyes~\cite{DeepEyes}        & 39.72 & 1155 & 2.67 & 14.89 \\
Mini-o3~\cite{mini_o3}         & 44.34 & 2291 & 5.29 & 8.38 \\
PixelEyes-4B-RL & 54.73 & 907  & 2.09 & 26.13 \\
\bottomrule
\end{tabular}%
}
\end{table}

\section{More Details of Pinpoint-Bench}
\noindent \textbf{Construction of Pinpoint-Bench.} 
The construction of Pinpoint-Bench consists of three progressive stages: raw data collection, collaborative annotation, and rigorous quality inspection. 
First, in the data collection stage, we crawled 1884 raw images from various internet videos and high-resolution image repositories. 
Second, during the collaborative annotation stage, we deployed a dedicated online platform where annotators worked concurrently. Out of the crawled pool, 721 images were successfully labeled; 471 images were flagged and discarded by the initial annotators as unsuitable (e.g., due to severe blurring or insufficient distractors) and automatically blacklisted from the task queue, while the remaining unassigned images (692) were directly deprecated. 
Finally, in the quality inspection stage, two senior reviewers cross-examined all annotated samples. They filtered out instances where questions were overly simplistic, lacked unique references, contained inaccurate mask groundings, or posed illogical premises. 
After eliminating 288 substandard samples during this final vetting process, we retained 433 high-quality examples to constitute the final benchmark.

\begin{figure}[!th]
  \centering
  \includegraphics[width=1.0\linewidth]{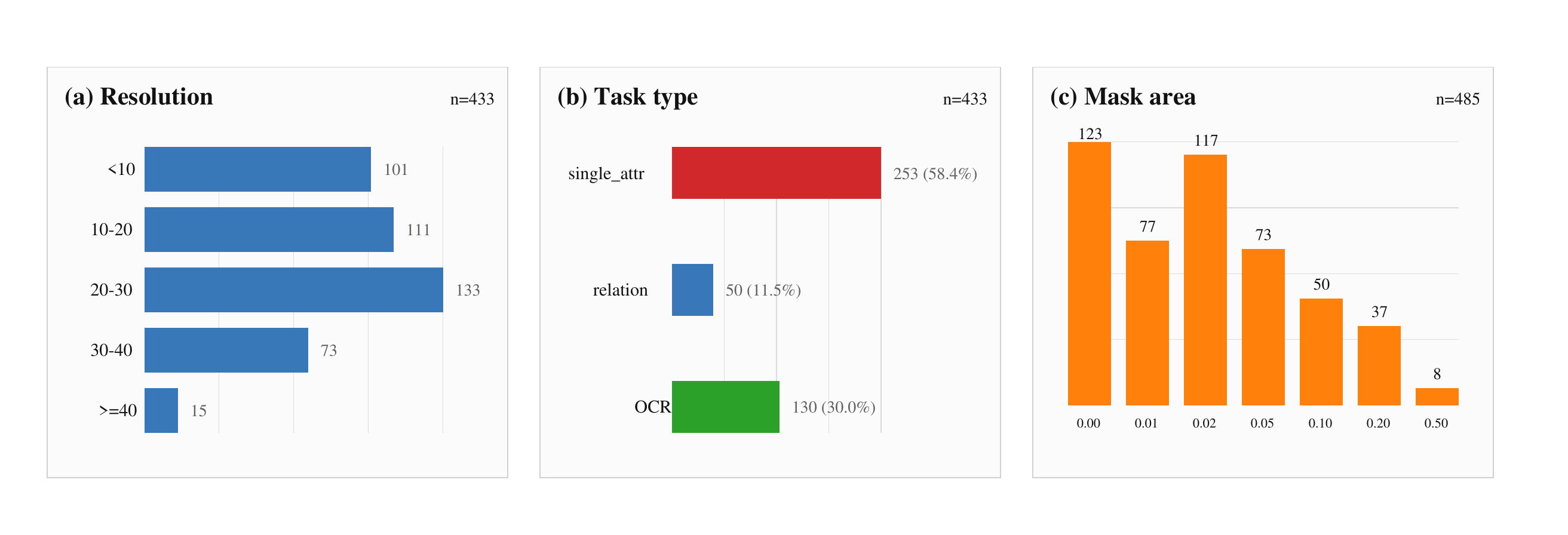}
  \caption{Statistical overview of Pinpoint-Bench. (a) Image resolution distribution, featuring ultra-high-definition samples with an average of $5500 \times 3516$ pixels. (b) Composition of task types, covering attribute recognition, spatial relationship reasoning, and OCR-centric queries. (c) Distribution of mask area ratios, where all targets
    occupy less than 1\% of the total pixels in the image, highlighting the "needle-in-a-haystack" nature of the benchmark.}
  \label{fig:statistics}
\end{figure}

\noindent \textbf{Statistics.} 
The tasks in Pinpoint-Bench encompass single-target attribute recognition, spatial relationship reasoning among two or three objects, and OCR-centric queries. As illustrated in Fig. \ref{fig:statistics}, images in our benchmark typically exceed 10 megapixels, with an average resolution of $5500 \times 3516$. 
While attribute recognition remains the primary focus, the targets are exceptionally challenging to locate: target masks occupy less than $1\%$ of the total image area, with a minuscule average footprint of only $0.07\%$, making Pinpoint-Bench an extremely rigorous evaluation suite.


\section{Limitations}
\noindent While PixelEyes demonstrates superior performance through its decoupled design, the current implementation introduces certain architectural complexity compared to fully end-to-end systems, leaving room for further optimization in computational efficiency.

\section{Visualizations}
\begin{figure}[!ht]
    \centering 
    \includegraphics[width=1\textwidth]{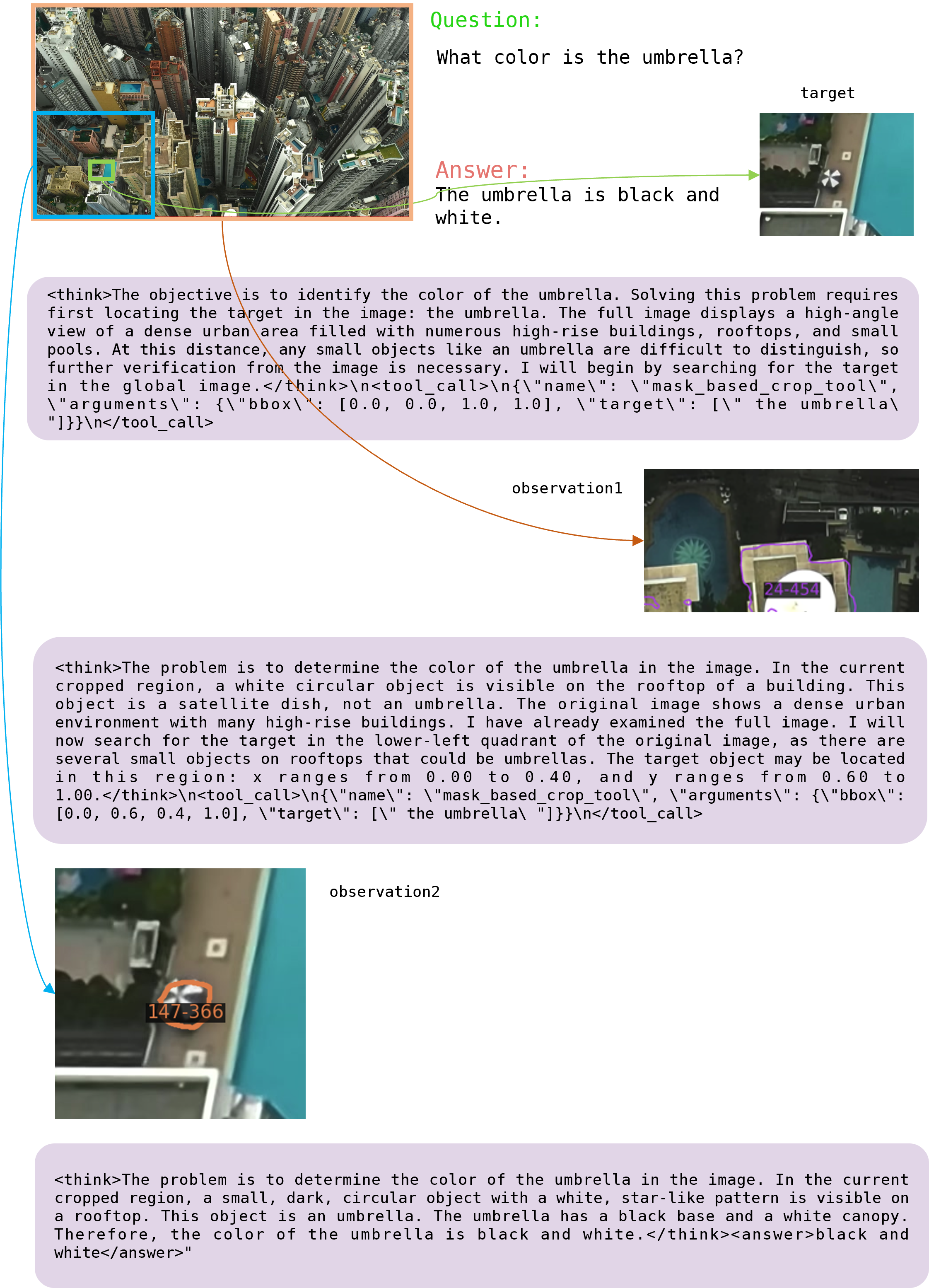}
    \caption{ 
\textbf{A hard case from our Pinpoint-Bench}. The question asks about the color of the umbrella. PixelEyes quickly grounded the target and answered the question \textbf{correctly}. The mask contours have been overlaid onto the original image for visualization.
}
    \label{fig:goodcase1}
\end{figure}

\begin{figure}[!ht]
    \centering 
    \includegraphics[width=0.9\textwidth]{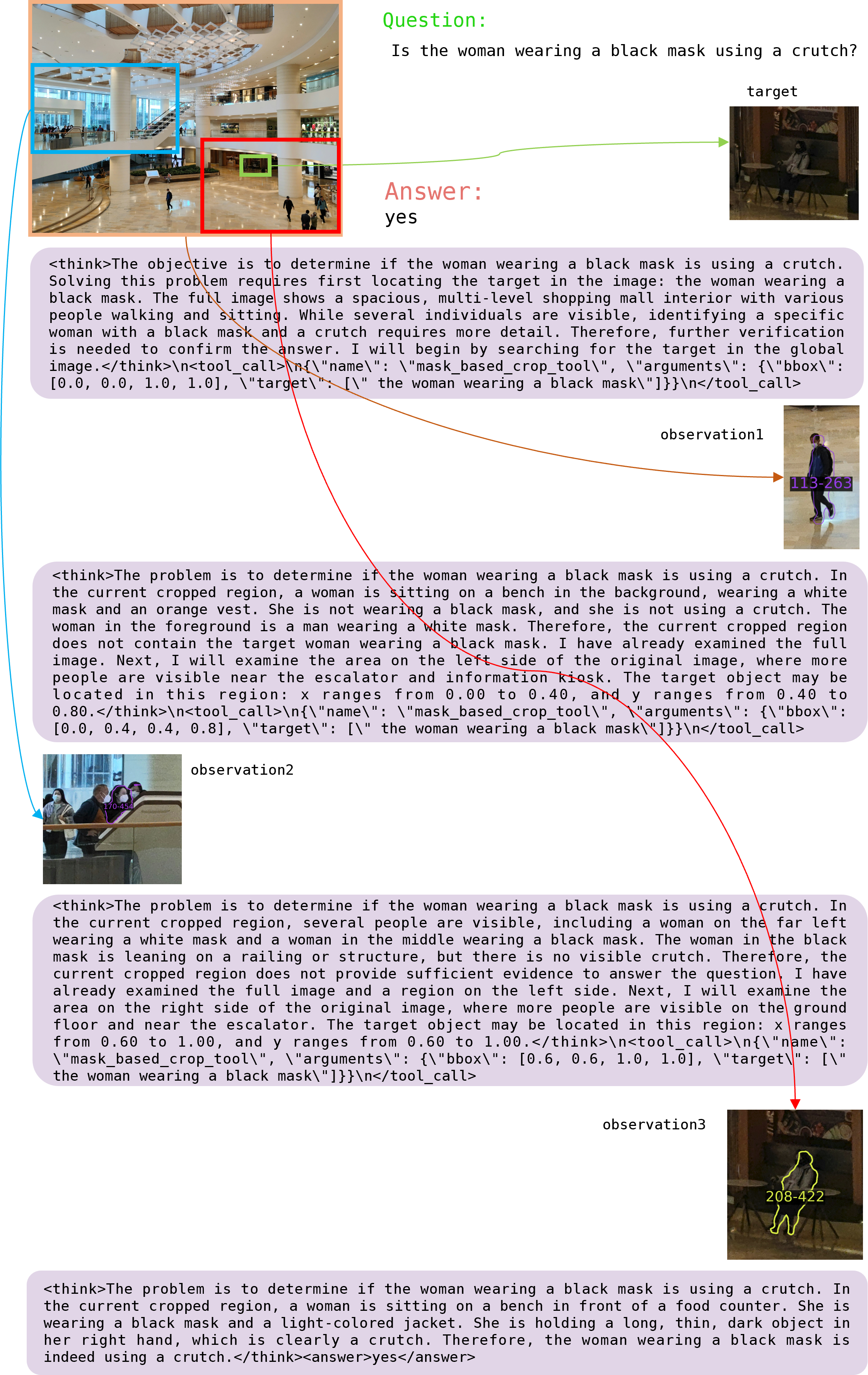}
    \caption{ 
\textbf{A hard case from our Pinpoint-Bench}. The question asks whether the woman wearing a black mask is using a crutch. By continuously proposing coarse bounding boxes and applying a mask grounder on them, PixelEyes progressively localized the target and ultimately answered the question \textbf{correctly}. The mask contours have been overlaid onto the original image for visualization.
}
    \label{fig:goodcase2}
\end{figure}

\begin{figure}[!ht]
    \centering 
    \includegraphics[width=1\textwidth]{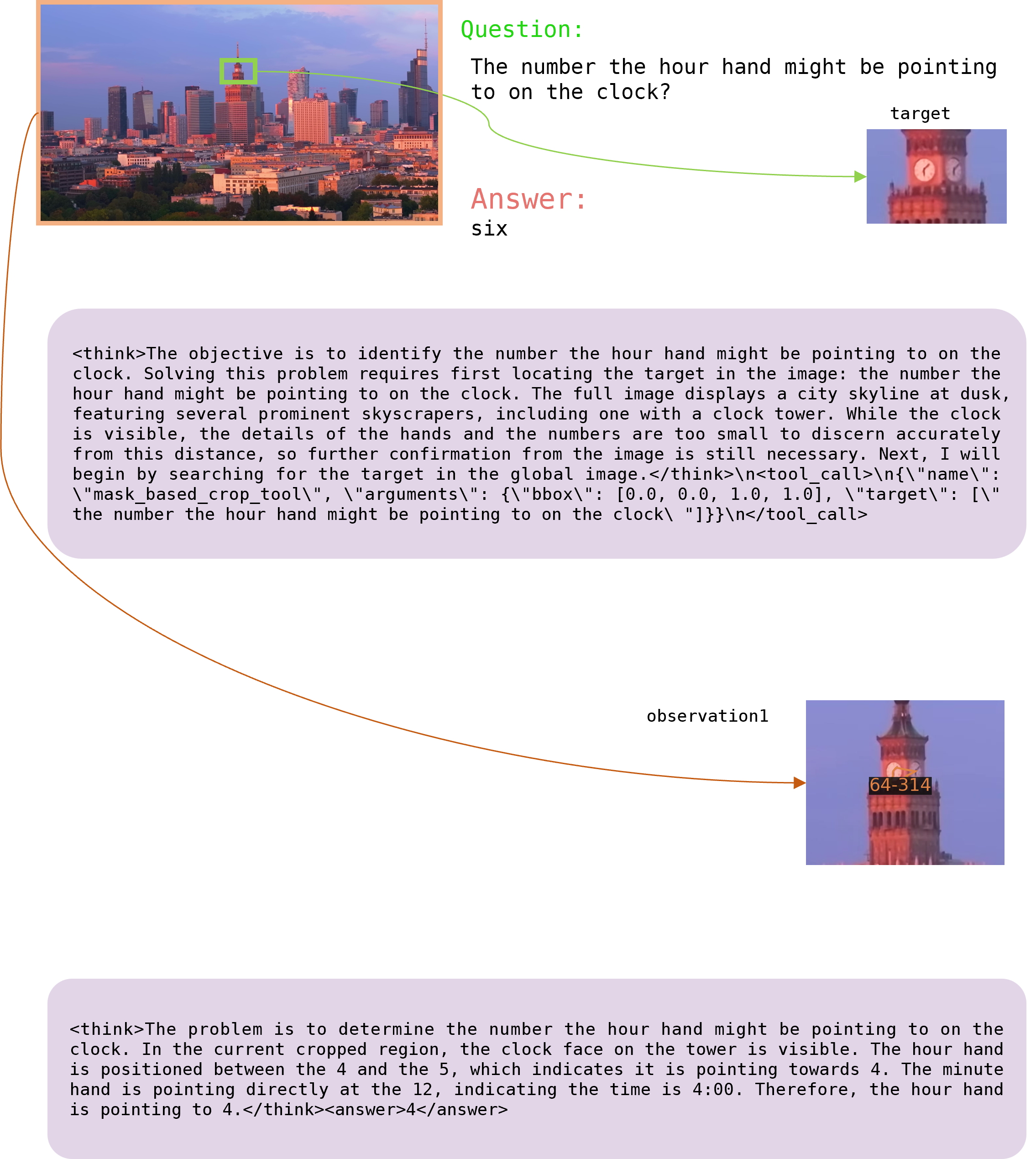}
    \caption{ 
\textbf{A hard case from our Pinpoint-Bench}. The question asks for the number the hour hand might be pointing to on the clock. Although PixelEyes quickly localized the target, it still answered \textbf{incorrectly}; however, this should be blamed on the base model itself. The mask contours have been overlaid onto the original image for visualization.
}
    \label{fig:hardcase1}
\end{figure}

\begin{figure}[!ht]
    \centering 
    \includegraphics[width=0.9\textwidth]{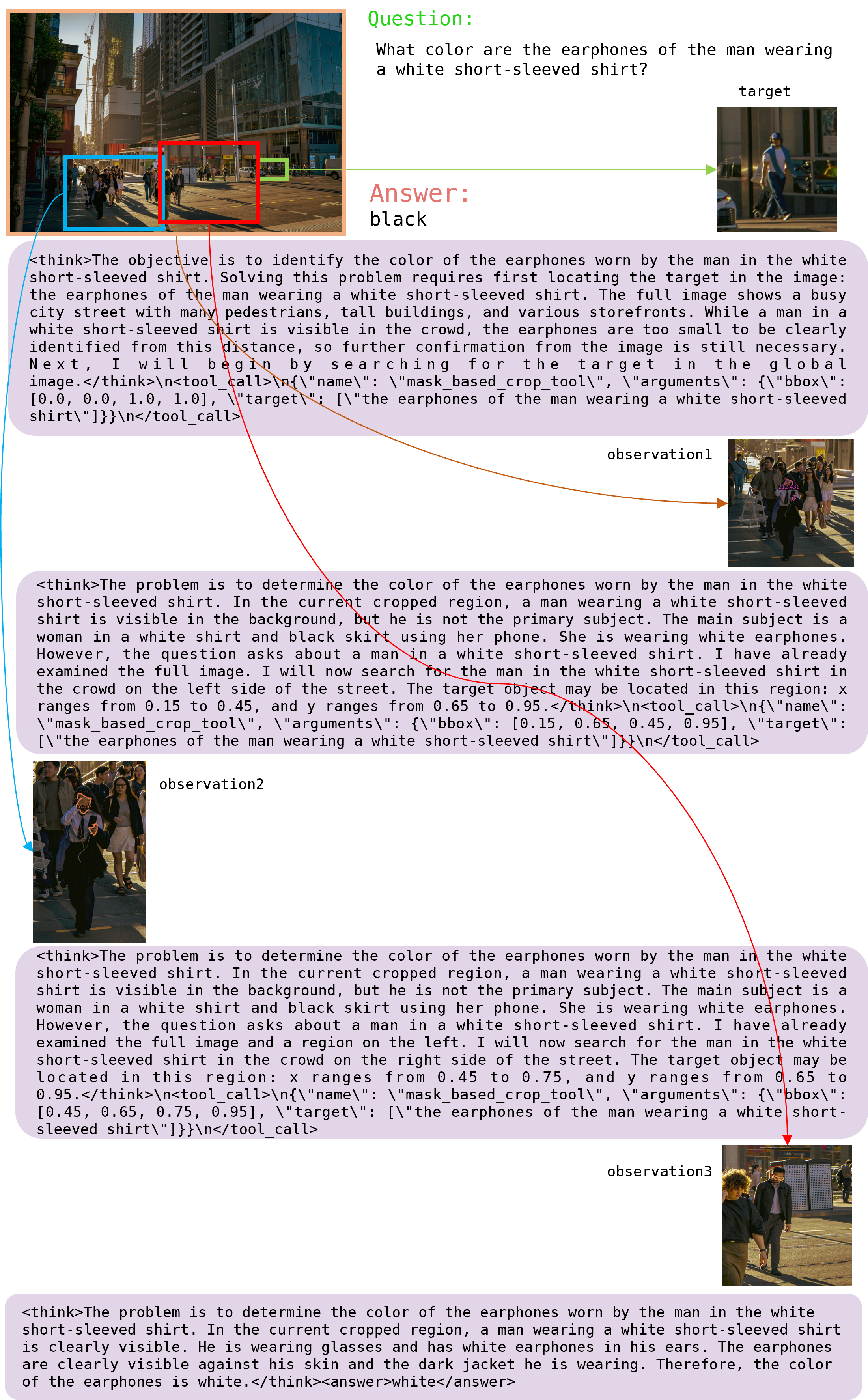}
    \caption{ 
\textbf{A hard case from our Pinpoint-Bench}. The question asks for the color of the earphones of the man wearing a white short-sleeved shirt. Despite significant lighting variations and multiple distractors (several people wearing earphones), PixelEyes repeatedly grounded a woman wearing earphones twice. It then confidently gave a \textbf{wrong} answer based on this faulty grounding, failing to consider the clothing constraint. Furthermore, it hallucinated, as the grounded man was not wearing earphones at all. The mask contours have been overlaid onto the original image for visualization.
}
    \label{fig:hardcase2}
\end{figure}

\noindent \textbf{Visualization of PixelEyes' Performance on Pinpoint-Bench.} 
As demonstrated by the two successful cases in Fig.~\ref{fig:goodcase1} and Fig.~\ref{fig:goodcase2}, PixelEyes exhibits an impressive capability to precisely localize microscopic targets alongside multi-turn tool-use search capacities. Conversely, the failure cases in Fig.~\ref{fig:hardcase1} and Fig.~\ref{fig:hardcase2} reveal that PixelEyes cannot circumvent the base model's inherent limitations, namely its deficient macro-search capability and hallucination vulnerabilities.

\begin{figure}[!ht]
    \centering 
    \includegraphics[width=1\textwidth]{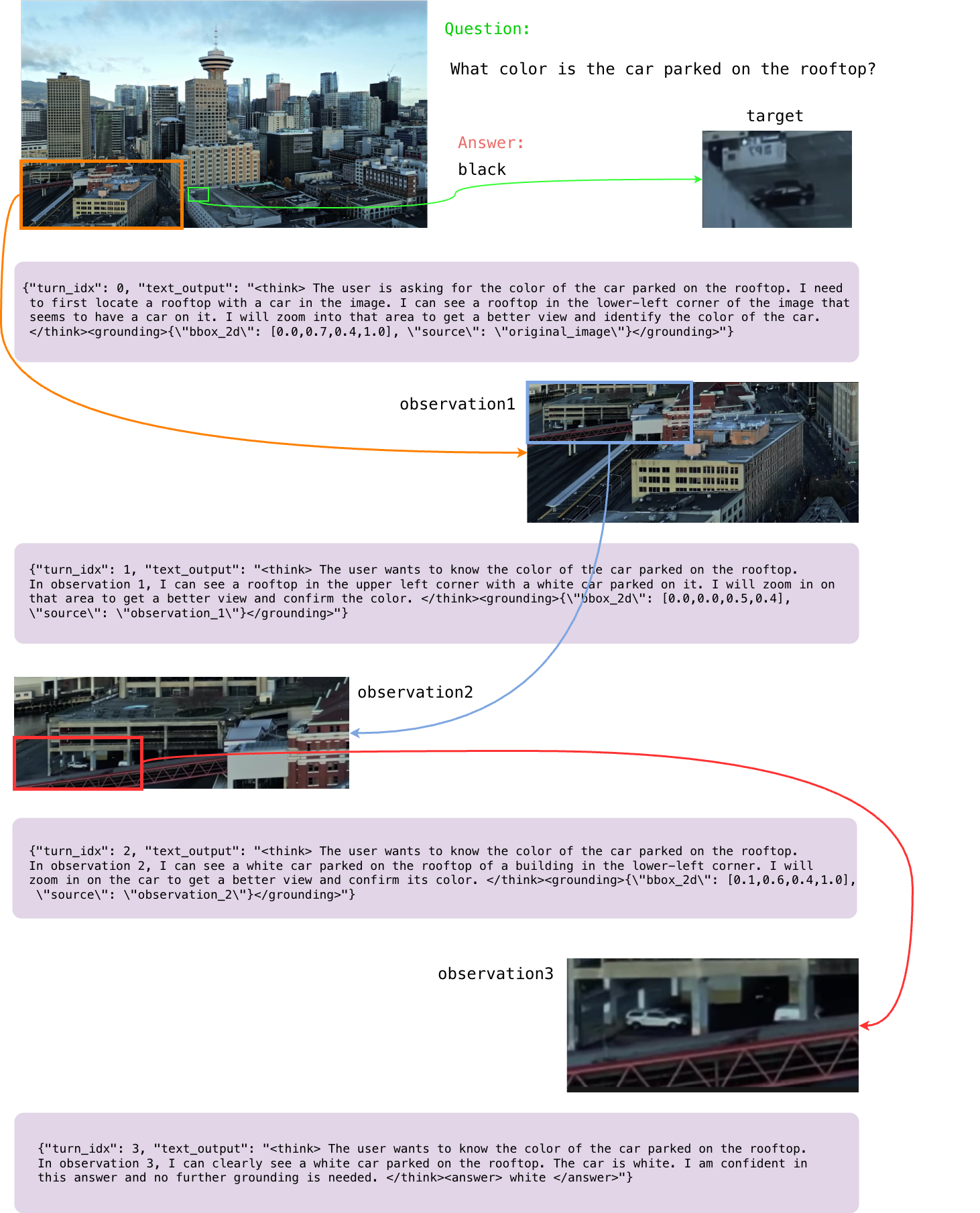}
    \caption{ 
\textbf{A hard case from our Pinpoint-Bench}. The question asks for the color of the car parked on the rooftop. During multi-turn cropping, Mini-o3 is distracted by a white car in an open parking structure and ultimately answers "white," mistaking the parking-garage car for the rooftop car.
}
    \label{fig:hardcase3}
\end{figure}

\begin{figure}[!ht]
    \centering 
    \includegraphics[width=1\textwidth]{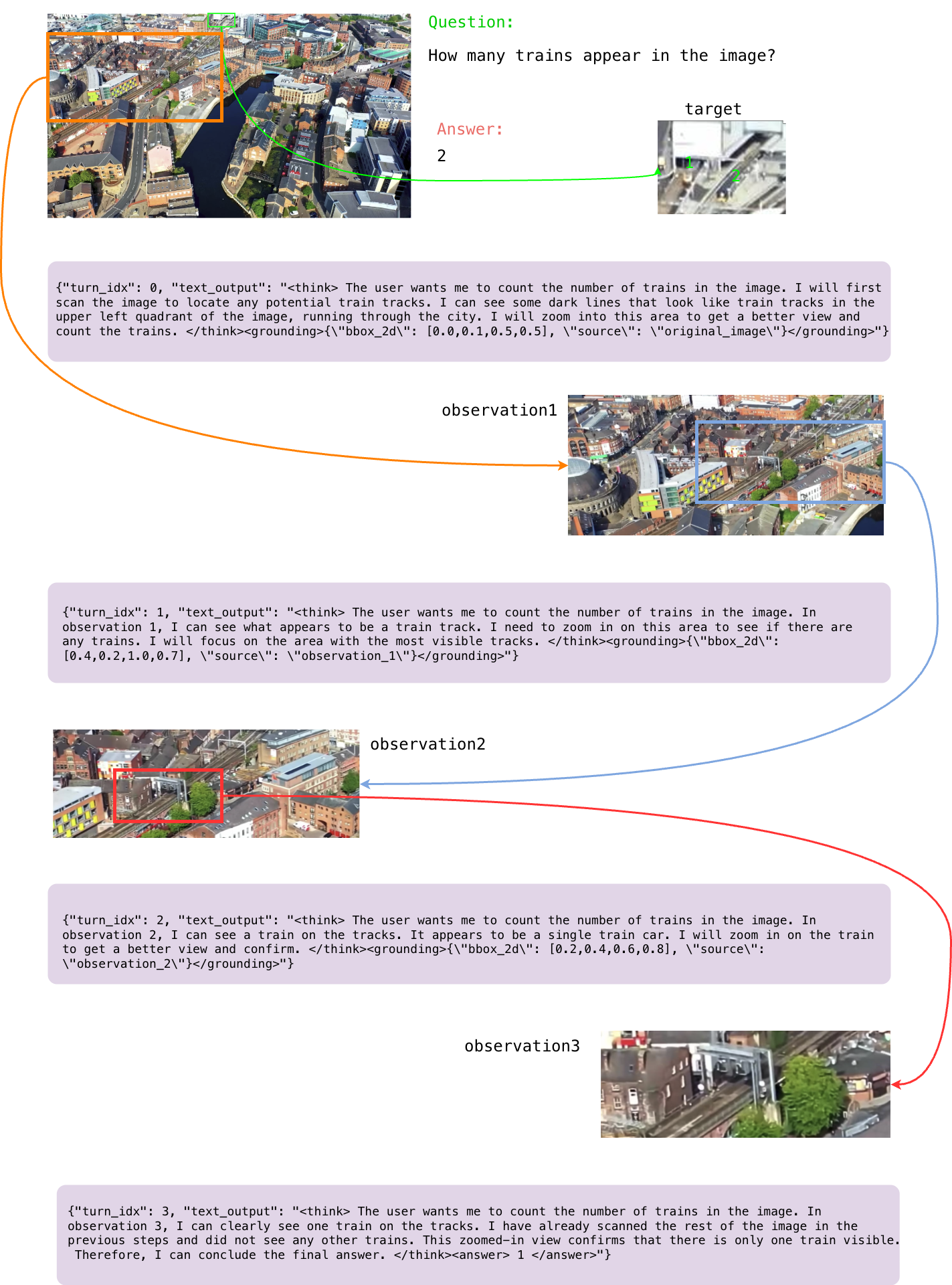}
    \caption{ 
\textbf{A hard case from our Pinpoint-Bench}. The question asks for the number of trains in the image. Through multi-turn cropping, Mini-o3 gives the wrong answer 1.
}
    \label{fig:hardcase4}
\end{figure}

\noindent \textbf{Challenging cases in Pinpoint-Bench.} 
To illustrate the difficulty of Pinpoint-Bench, Fig. ~\ref{fig:hardcase3} and Fig. ~\ref{fig:hardcase4} present two representative high-difficulty cases.

\end{document}